\PassOptionsToPackage{table}{xcolor}
\documentclass{article}
\usepackage[square,numbers]{natbib}

\usepackage[preprint]{neurips_2025}

\usepackage[utf8]{inputenc} % allow utf-8 input
\usepackage[T1]{fontenc}    % use 8-bit T1 fonts
\usepackage[pagebackref, breaklinks=true, colorlinks,citecolor=citecolor, linkcolor=linkcolor, bookmarks=false]{hyperref}
\usepackage{url}            % simple URL typesetting
\usepackage{booktabs}       % professional-quality tables
\usepackage{amsfonts}       % blackboard math symbols
\usepackage{nicefrac}       % compact symbols for 1/2, etc.
\usepackage{microtype}      % microtypography
\usepackage{xcolor}         % colors
\usepackage{tablefootnote}
\usepackage{colortbl}
\definecolor{citecolor}{HTML}{0071BC}
\definecolor{linkcolor}{HTML}{ED1C24}
\definecolor{datablue}{RGB}{67, 114, 194}
\usepackage{wrapfig}
\usepackage{graphicx}
\usepackage{pifont}
\usepackage{caption}
\usepackage{subcaption}
\usepackage{multirow} 
\usepackage{marvosym}

\title{HumanAesExpert: Advancing a Multi-Modality Foundation Model for Human Image Aesthetic Assessment}

\author{
    Zhichao Liao\textsuperscript{1\ \dag}
    \enspace Xiaokun Liu\textsuperscript{2}
    \enspace Wenyu Qin\textsuperscript{2}
    \enspace Qingyu Li\textsuperscript{2}
    \enspace Qiulin Wang\textsuperscript{2} 
    \\
    \enspace \textbf{Pengfei Wan}\textsuperscript{2}
    \enspace \textbf{Di Zhang}\textsuperscript{2} 
    \enspace \textbf{Long Zeng}\textsuperscript{1\ \Letter}
    \enspace \textbf{Pingfa Feng}\textsuperscript{1}
    \vspace{0.66em}
    \\\textsuperscript{1}Tsinghua University \quad\enspace \textsuperscript{2}Kuaishou Technology 
    \\{\tt\small liaozc23@mails.tsinghua.edu.cn} 
    \\{\tt\small zenglong@sz.tsinghua.edu.cn}\enspace{\tt\small fengpf@tsinghua.edu.cn} 
    \\{\tt\small \{liuxiaokun, qinwenyu, liqingyu, wangqiulin, wanpengfei, zhangdi08\}@kuaishou.com} 
    \vspace{0.66em}
    \\\url{https://humanaesexpert.github.io/HumanAesExpert/}
    \vspace{-1cm}
}

\begin{document}

\maketitle

{\let\thefootnote\relax\footnote{{\textsuperscript{\dag} Work done during internship at KwaiVGI, Kuaishou Technology.}}}
{\let\thefootnote\relax\footnote{{\textsuperscript{\Letter} Corresponding author.}}}

\begin{abstract}
Image Aesthetic Assessment (IAA) is a long-standing and challenging research task. However, its subset, Human Image Aesthetic Assessment (HIAA), has been scarcely explored.
To bridge this research gap, our work pioneers a holistic implementation framework tailored for HIAA. Specifically, we introduce \textbf{HumanBeauty}, the first dataset purpose-built for HIAA, which comprises 108k high-quality human images with manual annotations. To achieve comprehensive and fine-grained HIAA, 50K human images are manually collected through a rigorous curation process and annotated leveraging our trailblazing 12-dimensional aesthetic standard, while the remaining 58K with overall aesthetic labels are systematically filtered from public datasets. 
Based on the HumanBeauty database, we propose \textbf{HumanAesExpert}, a powerful Vision Language Model for aesthetic evaluation of human images. We innovatively design an Expert head to incorporate human knowledge of aesthetic sub-dimensions while jointly utilizing the Language Modeling (LM) and Regression heads. This approach empowers our model to achieve superior proficiency in both overall and fine-grained HIAA. Furthermore, we introduce a MetaVoter, which aggregates scores from all three heads, to effectively balance the capabilities of each head, thereby realizing improved assessment precision. Extensive experiments demonstrate that our HumanAesExpert models deliver significantly better performance in HIAA than other state-of-the-art models. 
\end{abstract}

\section{Introduction}
\begin{wrapfigure}{r}{0.42\textwidth}
  \centering
  % \vspace{-2em}
  \includegraphics[width=0.4\textwidth]{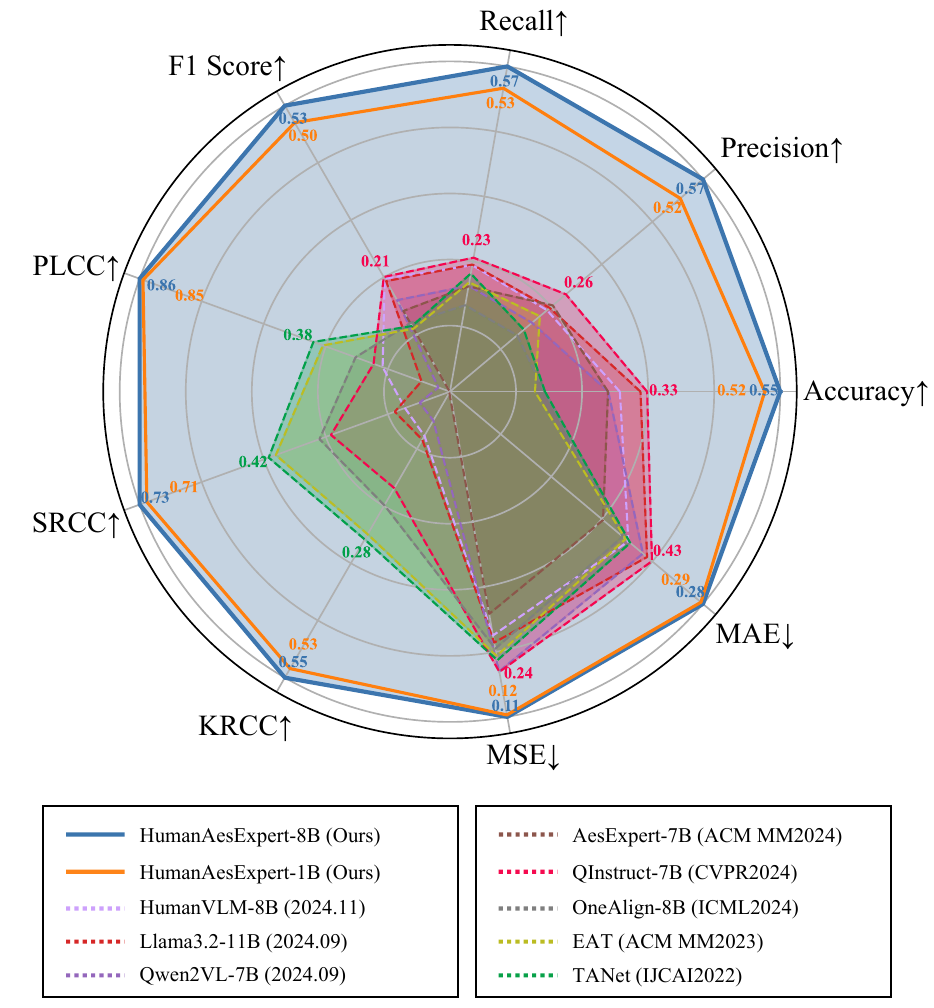}
  % \vspace{-12pt}
  \caption{\small 
  \textbf{Our HumanAesExpert, compared to existing state-of-the-art methods, shows exceptional improvements. }$\uparrow$ indicates that larger values are better, $\downarrow$ signifies the opposite.}
  \label{fig:leida}
  \vspace{-12pt}
\end{wrapfigure}

Human Image Aesthetic Assessment (HIAA) extends traditional Image Aesthetic Assessment (IAA) \cite{ava2012murray,deng2017image,yang2019comprehensive,zhang2021comprehensive,daryanavard2024deep} by shifting focus to quantitative evaluation of human-centric images. It aims to systematically analyze aesthetic dimensions (e.g., facial features, body shape,  environment) to assign measurable scores, transforming subjective human visual appeal into objective computational metrics. HIAA demonstrates extensive applicability across domains, such as employing quantitative aesthetic analysis to optimize content curation in social media recommendation systems, and implementing measurable quality metrics in generative AI workflows \cite{lin2024dreamfit,li2024anydressing,liao2024freehand,luo2024codeswap,luo2025object,wan2024grid,xian2025spf} to refine human-centered synthetic imagery.
However, HIAA has not been explicitly explored in previous work. Recently, benefiting from the outstanding performance of Vision Language Models (VLMs) \cite{qwen2vl2024wang,llava152024liu,internvl22024chen,liu2024deepseek,guo2025deepseek} in multimodal tasks, VLM-based IAA methods \cite{qalign2023wu,qinstruct2024wu,aesexpert2024huang,uniaa2024zhou} have secured significant breakthroughs.
Nevertheless, as shown in Fig.~\ref{fig:leida}, general IAA approaches exhibit suboptimal performance when handling specialized HIAA tasks, severely restricting the practical application of HIAA. 
% in real-world scenarios.
Consequently, it is urgent to develop a holistic framework capable of conducting systematic and professional aesthetic assessment for human-centric images to bridge these critical research gaps.

Diverse and informative human image data with aesthetic labels are essential for HIAA. 
However, to our knowledge, there is no open-source HIAA dataset in existence.
To address the data gap, we introduce \textbf{HumanBeauty}, the first dataset dedicated to HIAA research.
We first employ face detection algorithms \cite{retinaface2020deng} to curate 58,564 human images from public IAA datasets \cite{ava2012murray, scut55002018liang, mebeauty2022lebedeva, BAID2023yi, agiqa2023li, TAD66k2022he} and unify their aesthetic score scales.
Notably, these data exhibit two significant shortcomings in achieving comprehensive and fine-grained HIAA:
\textbf{1) Image level:} insufficient human images with poor coverage of HIAA attributes (e.g., numerous images contain only facial or head regions, neglecting full-body aesthetics).
\textbf{2) Annotation level:} Only overall aesthetic scores, with the absence of sub-dimensional granular annotations supported by a systematic aesthetic standard.
To resolve these issues, guided by extensive consultations with aesthetic experts and studio professionals, we establish a 12-dimensional HIAA standard which encompasses all critical attributes of HIAA.
Based on this standard, we implement an iterative training-testing protocol to train annotators.
Ultimately, we organize 368 certified volunteers to collect 50,022 informative human images from the Internet, with each image undergoing rigorous multi-dimensional manual annotation following our aesthetic standard.
Integrating data from both phases, we build our HumanBeauty dataset, which contains a total of 108k annotated human images and is the largest HIAA dataset compared to existing related datasets, as shown in Tab.~\ref{tab:dataset-compare}.

\begin{wraptable}{r}{0.5\textwidth}
    \centering
    \vspace{-12pt}
    \caption{\small \textbf{Comparisons with existing related datasets.}}%
    % \renewcommand\arraystretch{1}
    % \vspace{+5pt}
    \resizebox{1\linewidth}{!}{%
    \begin{tabular}{c|ccc}
    \toprule
    Dataset                  & Publication Year & \textit{\textcolor{datablue}{Human}} / Total & Theme              \\ \midrule
    Gray et al. \cite{predictingfacial2010gray}              & 2010             & \textcolor{datablue}{\textit{2,056}} / 2,056              & Face               \\
        AVA  \cite{ava2012murray}                     & 2012             & \textcolor{datablue}{\textit{44,920}} / 255,530            & General            \\
    SCUT-FBP 5500 \cite{scut55002018liang}           & 2018             & \textcolor{datablue}{\textit{5,500}} / 5,500              & Face               \\
    MEBeauty \cite{mebeauty2022lebedeva}                & 2022             & \textcolor{datablue}{\textit{2,550}} / 2,550              & Face               \\
    ICAA17K \cite{ICAA17K2023he}                 & 2023             & \textcolor{datablue}{\textit{3894}} / 17,726             & Color              \\
    BAID \cite{BAID2023yi}                    & 2023             & \textcolor{datablue}{\textit{13,869}} / 60,337             & Artistic                \\
    AGIQA \cite{agiqa2023li}                   & 2023             & \textcolor{datablue}{\textit{1218}} / 2,982              & AI-Generated \\
    AesBench \cite{aesbench2024huang}                 & 2024             & \textcolor{datablue}{\textit{894}} / 2,800              & General            \\
    AesMMIT \cite{aesexpert2024huang}               & 2024             & \textcolor{datablue}{\textit{5179}} / 21,904             & General            \\ \midrule
    \textbf{HumanBeauty (Ours)} & \textbf{2025}    &  \textit{\textbf{\textcolor{datablue}{108,586}}} / 108,586     & \textbf{Human}     \\ \bottomrule
    \end{tabular}
    }
    \label{tab:dataset-compare}
    \vspace{-11pt}
\end{wraptable}

Current VLM-based IAA methods have demonstrated promising efficacy, they predominantly adopt two architectural paradigms.
% as shown in Fig.~\ref{fig:comparison_model}:
(a) LM head-based VLMs: Methods \cite{lift2024wang,qalign2023wu,uniaa2024zhou,qinstruct2024wu} discretize continuous scores into textual labels (e.g., ``good'' and ``bad'') for training, and then map predictions back to scores via softmax probabilities during inference. However, supervision with discrete labels hinders the accurate fitting of continuous scores. 
(b) Regression head-based VLMs: Approaches \cite{rich2024liang,videoscore2024he} replace LM heads with Regression heads to predict Mean Opinion Scores (MOS) directly. While mitigating discretization errors, it sacrifices the text comprehension proficiency of large language models (LLMs), reducing interpretability. Moreover, learning multi-dimensional aesthetic scores via a single Regression head risks distribution conflicts and domain-specific accuracy degradation. 
Therefore, to simultaneously enable granular aesthetic evaluation and preserve the model's text comprehension capabilities with scoring precision, we propose \textbf{HumanAesExpert}, which introduces an Expert head to integrate fine-grained aesthetic knowledge, working in tandem with the LM head and Regression head.
Furthermore, we design a MetaVoter that aggregates scores from three heads to generate the final scores, which employs learnable weights to balance the contributions of each head, improving the aesthetic assessment accuracy.
Finally, we conduct extensive experiments to demonstrate that our approach achieves state-of-the-art performance (SOTA) in HIAA compared to previous works. In summary, our contributions are as follows: \textbf{1) }To our knowledge, we are the pioneering work for HIAA and introduce the \textbf{HumanBeauty} dataset, which is the first large-scale dataset dedicated to HIAA, comprising 108K manually annotated human images. \textbf{2) }We propose the \textbf{HumanAesExpert}, a foundation VLM for HIAA, which innovatively introduces an Expert Head and MetaVoter to achieve fine-grained aesthetic evaluation and balance contributions of multi heads. \textbf{3) }Extensive experiments demonstrate that our models achieve SOTA performance in HIAA across all metrics. \textbf{4) }We publicly release our datasets, models, and codes to drive the development of the HIAA community.

\section{Related Work}
\noindent{\textbf{Image Aesthetic Assessment Datasets. }}The most popular and largest general IAA dataset is the AVA dataset \cite{ava2012murray}. Over the years, AVA has significantly advanced the development of the IAA community. It has also served as the source dataset for several subsequent datasets, including recent works like AesBench \cite{aesbench2024huang} and AesExpert \cite{aesexpert2024huang}. However, general IAA often lacks clear scoring criteria, leading to discrepancies in aesthetic evaluation points that human raters focus on. This results in abnormal score distributions and data quality decline. 
Given the extensive application of facial attractiveness in psychology, datasets \cite{predictingfacial2010gray,scut55002018liang,mebeauty2022lebedeva} using facial beauty as a scoring criterion have been continuously developed. While ICAA17K \cite{ICAA17K2023he} focuses on image color, BAID \cite{BAID2023yi} on artistic IAA, and AGIQA \cite{agiqa2023li} on AI-generated images, as summarized in Tab.~\ref{tab:dataset-compare}, there remains a lack of open-source datasets centered on human subjects. Such datasets should consider not only facial beauty but also factors like body shape, environment, and more.

\noindent{\textbf{Traditional Assessment Methods. }}Traditional assessment methods involve using Convolutional Neural Networks (CNN) \cite{resnet2016he,mobilenetv22018sandler,alexnet2012krizhevsky,resnext2017xie,vggnet2014simonyan} or Vision Transformers (ViT) \cite{vit2020dosovitskiy,swintransformer2021liu,mae2022he,clip2021radford} to extract image features and predict scores using a Regression head. CNN-based methods \cite{mlsp2019hosu, TAD66k2022he}, pre-trained on ImageNet \cite{imagenet2009deng}, have been extensively studied. However, these methods often lose the prior knowledge of pre-training in the fine-tuning process, limiting the models' ability to understand aesthetics and effectively identify salient regions. In addition, achieving human-level aesthetic alignment requires priors comparable to humans'. CNNs and ViTs pre-trained on ImageNet fall short of this objective, and there remains a lack of causal reasoning. The models do not understand which factors contribute to the overall aesthetic scores.

\noindent{\textbf{VLM-based Assessment Methods. }}Compared to traditional assessment methods, VLMs like GPT-4 \cite{achiam2023gpt} exhibit strong causal reasoning capabilities. Recent works, such as Q-Bench \cite{qbench2023wu}, Q-Align \cite{qalign2023wu}, and UNIAA-LLaVA \cite{uniaa2024zhou}, propose mapping scores to discrete text-defined levels, which are used to fine-tune VLMs. During inference, the softmax function is applied to the logits of the rating-level words corresponding to the score token to derive scores. However, this approach fails to leverage the continuous supervisory signal from the scores. To better utilize this signal, works \cite{rich2024liang,videoscore2024he} replace the LM head with a Regression head. However, the LM head is a critical component for causal reasoning in LLMs, and removing it leads to similar issues observed in traditional assessment methods \cite{lift2024wang}. In addition, this method struggles with cross-dimension data. Specifically, scores from different domains are learned through the same Regression head, resulting in information coupling that hinders proper fitting for domain-specific tasks. 
% To solve this, we propose using the expert head to learn domain-specific scores. The MetaVoter is then employed to balance the outputs of multiple heads, collectively determining the final score.

\begin{figure*}[tb]
  \centering
  \includegraphics[width=1\textwidth]{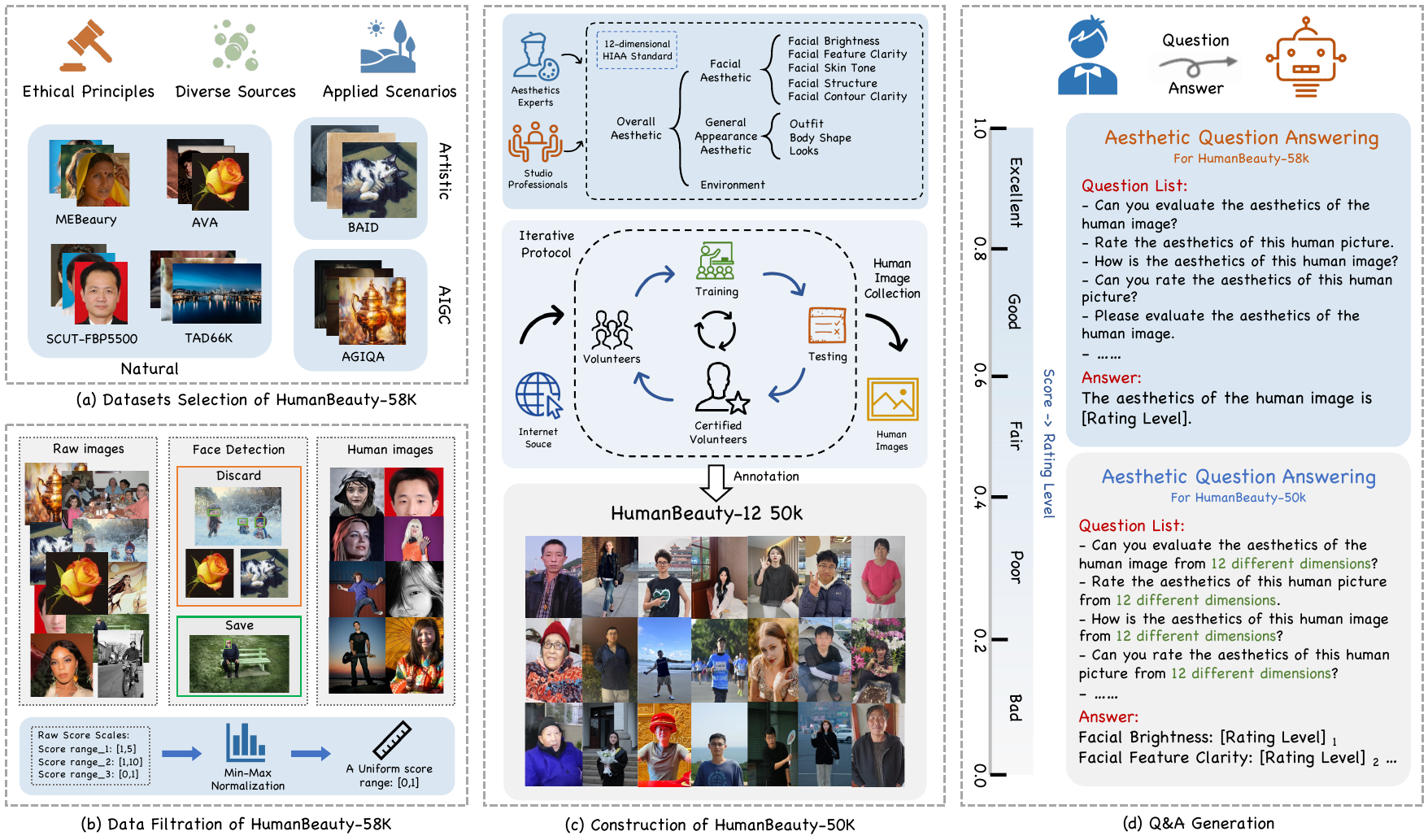}
  \caption{ \textbf{HumanBeauty construction pipeline}. First, we select six diverse open-source datasets as data sources and perform data filtering to build our HumanBeauty-58k. Additionally, we manually collect and annotate 50k human images across multiple dimensions to create our HumanBeauty-50k. Finally, we map all the scores into text of rating level to form QA pairs for training.
  }
  \label{fig:data-pipeline}
\end{figure*}

\section{HumanBeauty Dataset Construction}
% Our HumanBeauty Dataset is composed of two segments: HumanBeauty-58k and HumanBeauty-50k. In this section, we will detail the construction process of each segment.
\label{sec:HumanBeauty}

\subsection{HumanBeauty-58K}
Although existing IAA datasets cannot completely meet our task requirements, they include a moderate number of human images with overall aesthetic scores, which are capable of improving the model's overall assessment ability, robustness, and dataset diversity.
Following previous works \cite{aesexpert2024huang, qbench2023wu}, we collect 58k relevant human images from public datasets.

\noindent \textbf{Datasets Selection.} Firstly, adhering to ethical principles, we only consider open-source datasets free of gore, violence, and other harmful content.
Secondly, as IAA presents cross-domain and out-of-domain challenges, it is vital to incorporate diverse types of human images, such as natural, artistic, and AI-generated images.
Thirdly, given the application distribution of downstream tasks, we primarily focus on natural human images.
Motivated by the above considerations, we ultimately select SCUT-FBP 5500 \cite{scut55002018liang}, MEBeauty \cite{mebeauty2022lebedeva}, AVA \cite{ava2012murray}, TAD66K \cite{TAD66k2022he}, BAID \cite{BAID2023yi}, and AGIQA \cite{agiqa2023li} as our data sources, shown in Fig.~\ref{fig:data-pipeline} (a). 
% These datasets are open-source, ethically compliant, and include a variety of image types, with a higher proportion of natural human images.

\noindent \textbf{Data Filtration.} 
Due to the significant impact of facial aesthetics on HIAA, it necessitates that human images cover facial regions as thoroughly as possible. Similar to LAION-FACE \cite{laionface2022zheng}, FLIP-80M \cite{flip2024li} and HumanVLM \cite{humanvlm2024dai}, we also use RetinaFace \cite{retinaface2020deng}, a face detection algorithm, to filter human-related data from general datasets, shown in Fig.~\ref{fig:data-pipeline} (b).
Since single-human IAA underpins multi-human interaction analysis, we curate our dataset to include only single-face images, ensuring methodological continuity for future multi-human studies.
Eventually, we collect 58,564 human images, which are suitable for our task.

\noindent \textbf{Normalization. }Owing to varying aesthetic score scales, we independently normalize scores from each dataset to [0,1] using Min-Max normalization, as defined by the following equation:
\begin{equation}
    y_i=\frac{x_i-\min(x)}{\max(x)-\min(x)} \ ,
\end{equation}
\noindent where $x$ is a set of filtered dataset scores, and $y$ is the normalized score.

\subsection{HumanBeauty-50K}
To address the limitations of \textbf{Annotation level} and \textbf{Image level} in open-source data for achieving comprehensive and fine-grained HIAA, we additionally collect 50k high-quality human images with nuanced 12-dimensional aesthetic annotations. The process is shown in Fig.~\ref{fig:data-pipeline} (c).

\noindent \textbf{12-dimensional HIAA Standard. }Due to the inherently subjective and vague nature of aesthetics, individuals exhibit significant differences in their evaluations and often lack concrete criteria to justify their judgments.
Motivated by the desire to reconcile aesthetic disparities among individuals and establish a systematic and attributable assessment standard, we collect extensive opinions from aesthetics experts and studio professionals, summarized as follows:
i) The standard should cover all critical aspects that affect HIAA. ii) The standard should be hierarchical and attributable, with the attributes of the smallest sub-dimensions being decoupled from each other and directly contributing to their parent dimensions.
iii) The facial region has a significant impact on HIAA and requires more fine-grained evaluation dimensions.
iv) The general appearance and environment are equally important for the overall aesthetics.
Based on these, we establish the 12-dimensional aesthetic assessment standard for HIAA, which is similar to the sub-dimensional decomposition methodology in IAA, as illustrated in Fig.~\ref{fig:data-pipeline} (c).
The standard includes three groups: \textbf{1) Facial aesthetic: facial brightness, facial feature clarity, facial skin tone, facial structure, and facial contour clarity. \textbf{2)} General appearance aesthetic: outfit, body shape, and looks. 
\textbf{3)} Environment.} These sub-dimensions simultaneously contribute to the overall aesthetic.

\noindent \textbf{Iterative Protocol. } Our dataset construction is highly professional and challenging. 
To ensure data quality, we design a rigorous Iterative Training-Testing Protocol to qualify our volunteers. 
Specifically, we invite 455 volunteers from various fields, and aesthetic experts train them on human image collection principles and scoring criteria based on the 12-dimensional HIAA standard.
Volunteers undergo qualification tests, with those failing to meet requirements exiting the project, while the qualified proceed to subsequent phases of training and testing in an iterative manner. 
Ultimately, 368 certified volunteers are retained to collect and annotate human images for our dataset.
Notably, we periodically repeat this protocol during the formal image collection and annotation process to ensure a high-quality dataset.

\noindent \textbf{Human Image Collection. } We organize the certified volunteers to collect human images from the Internet, following the principles below:
\textbf{1)} Exclusion of ethical violations (e.g., privacy, explicit or violent content). 
Note that all non-public individual images in the manuscript have been authorized.
\textbf{2)} Images should focus on the face while including body representation of the subject (excluding portraits that focus solely on the face/head).
\textbf{3)} Images could contain multiple individuals, but must feature a single dominant human subject.
% After the initial image collection, we manually balance the dataset to eliminate data bias, ensuring an even distribution of key factors such as ethnicity and gender.
Ultimately, we obtain 50,022 human images.

% 经过初步的图片收集后，我们为了消除数据bias，我们人为的进行了数据balance，均衡了数据集中 ethnicity, gender, 等关键因素

\noindent \textbf{Aesthetic Annotation. } We organize the certified volunteers to annotate the collected human images across 12 dimensions.
Each dimension is scored within the range [0,1], and each image is annotated by at least 9 raters. The final score for each dimension is the Mean Opinion score (MOS), calculated as the average of all raters' scores.
We provide detailed evaluation criteria and scoring guidelines for each sub-dimension in the \textit{Appendix}.

\subsection{Question Answer Generation}
To meet the requirements of QA training pairs in VLMs and rating-level text supervision for the LM head \cite{qalign2023wu}, we convert the scores in existing datasets into discrete rating-level text. 
Inspired by Q-Align \cite{qalign2023wu} and UNIAA \cite{uniaa2024zhou}, we divide the scores into five equal intervals and map them to discrete text labels using a piecewise function:
\begin{equation}
\label{eq:1}
    Z(s)=z_i\mathrm{~if~}\frac{z-1}{5}<s\leq \frac{z}{5},
\end{equation}
\noindent where $\{z_i|_{z=1}^5\}=\{bad,poor,fair,good,excellent\}$, defined by ITU \cite{series2012methodology}, are standard text of rating levels. 
As shown in Fig.~\ref{fig:data-pipeline} (d), for HumanBeauty-58K, we randomly select a question from a group of paraphrases \cite{qalign2023wu} and directly map the overall aesthetics score to rating level as the answer, to construct QA pair, e.g.:
\begin{itemize}
    \small
    \item \textit{\textbf{Question: Rate the aesthetics of this human picture.}}
    \item \textit{\textbf{Answer: The aesthetics of the image is [Rating Level].}}
\end{itemize}
For HumanBeauty-50K, we also select a question about 12-dimensional HIAA from a set of conditional paraphrases and map the scores of each sub-dimension to rating levels, assembling the answer in the format: [Sub-dimension Name]: [Rating level]$_{n}$, where $n \in [1,12]$. For example:
\begin{itemize}
    \small
    \item \textit{\textbf{Question: Can you evaluate the aesthetics of the human image from 12 different dimensions?}}
    \item \textit{\textbf{Answer: Facial Brightness: [Rating level]$_{1}$}} \\
    \textit{\textbf{Facial Feature Clarity: [Rating Level]$_{2}$     ...} .}
\end{itemize}

\section{HumanAesExpert Model}
\label{sec:HumanAesExpert}

% \begin{figure}[tb]
%     \centering
%     \begin{subfigure}[b]{0.45\textwidth}
%         \centering
%         \includegraphics[width=\textwidth]{figures/model-overview.pdf}
%         \caption{Overview of HumanAesExpert.}
%     \end{subfigure}
%     \vfill
%     \begin{subfigure}[b]{0.45\textwidth}
%         \centering
%         \includegraphics[width=\textwidth]{figures/ExpertHead.pdf}
%         \caption{The structure of the Expert Head.}
%     \end{subfigure}
%     \caption{(a) The training path of the human images with only overall annotations and 12-dimensional annotations are highlighted with purple and yellow, respectively. 
%      (b) The Expert head is a sparsely connected MLP, with each node being supervised.}
%     \label{fig:model-overview}
% \end{figure}

In this section, we introduce the \textbf{HumanAesExpert} model. It employs both the LM head and the Regression head to leverage their respective strengths: text comprehension and continuous score learning.
However, a single Regression head is unable to learn scores from different dimensions across domains, like the overall scores from HumanBeauty-58K and the 12-dimensional scores from HumanBeauty-50K. 
Additionally, it cannot reflect the hierarchical structure of our 12-dimensional aesthetic assessment standard. To address this issue, we innovatively introduce the Expert head to integrate the knowledge of aesthetic sub-dimension, achieving fine-grained HIAA. Moreover, we propose a MetaVoter to combine the capabilities of these heads. This approach allows the three heads to collaboratively determine the final score. 
Next, we will introduce each of these components individually. Given an image $x$ and a text prompt $p$, we can obtain the final layer output $h=M(x,p)$ of the VLM model $M$, which will be fed to different heads.

\noindent \textbf{LM Head. }The LM head $H_{LM}$ \cite{qalign2023wu,uniaa2024zhou}, typically a linear layer, is used to obtain the logits $l=H_{LM}(h)$ for all words in the vocabulary. During training, these logits are used to compute the cross-entropy loss with the ground truth:
\begin{equation}
    \mathcal{L}_{Entropy}=-\frac{1}{N}\sum_{n=1}^N\sum_{i=1}^L y_{(i,n)}\log(l_{(i,n)}),
\end{equation}
\noindent where $N$ is the length of the ground truth answer, $L$ is the length of the vocabulary, and $y\in\{0,1\}$. During inference, the logits $p$ of the $\{bad, poor, fair, good, excellent\}$ are extracted from the logits $l$, and the weighted softmax function is applied to get a numerical score. The formula is as follows:
\begin{equation}
S_{LM}=\sum_{i=1}^5\frac{i\cdot e^p}{\sum_{j=1}^5e^p}.
\end{equation}

\begin{wrapfigure}{r}{0.55\textwidth}
  \centering
    \vspace{-10pt}
    \begin{subfigure}[b]{0.55\textwidth}
        \centering
        \includegraphics[width=\textwidth]{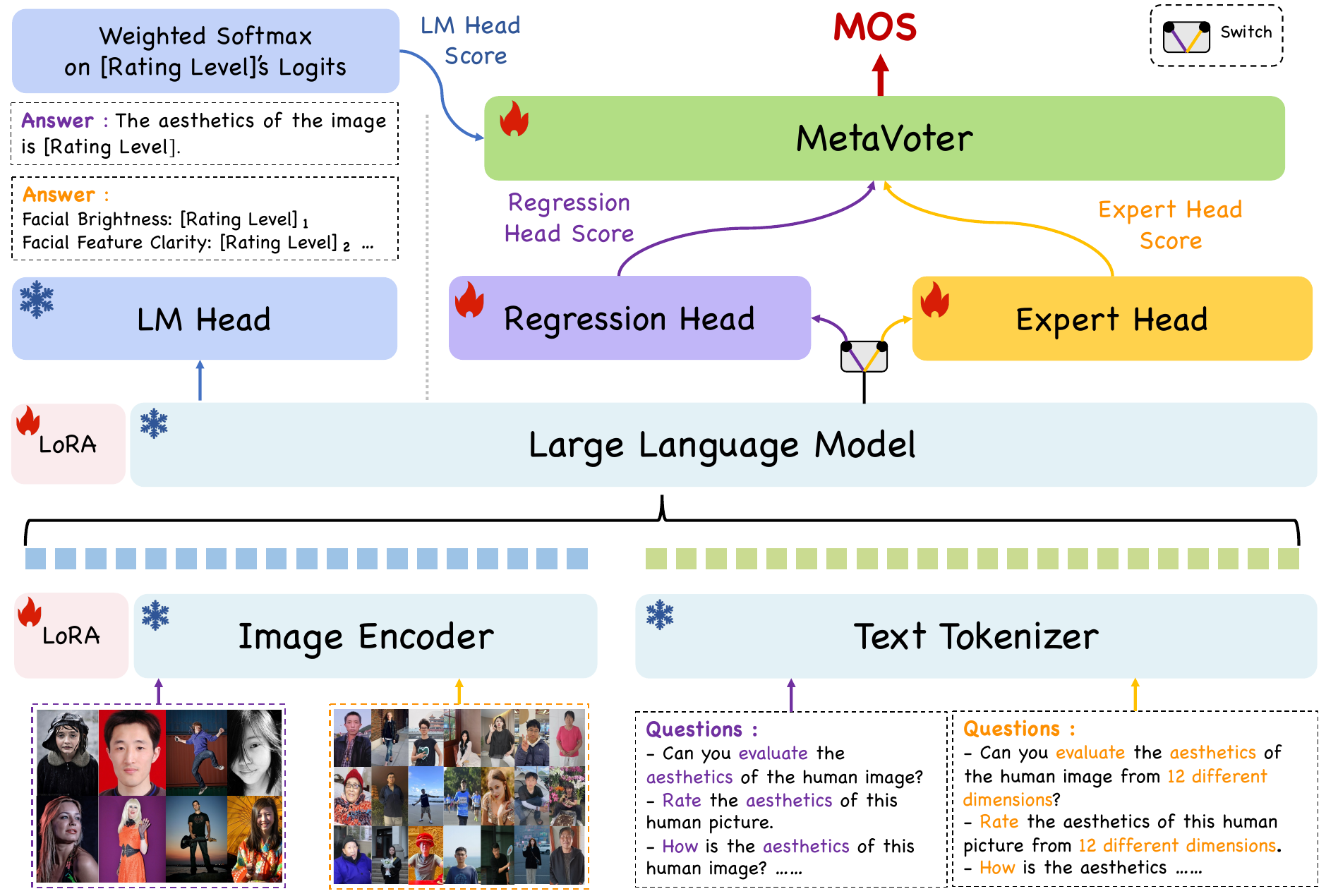}
        \caption{Overview of HumanAesExpert.}
    \end{subfigure}
    \vfill
    % \vspace{+2pt}
    \begin{subfigure}[b]{0.55\textwidth}
        \centering
        \includegraphics[width=\textwidth]{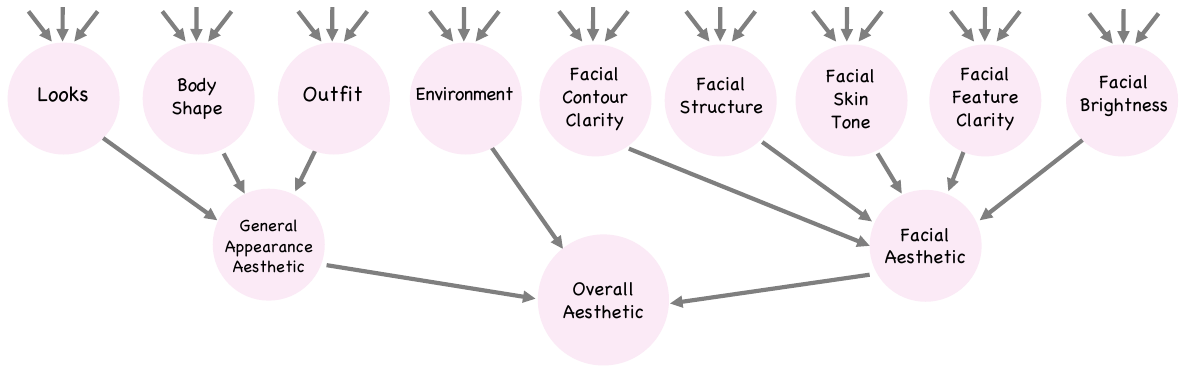}
        \caption{The structure of the Expert Head.}
    \end{subfigure}
    \caption{\small (a) The training path of the human images with only overall annotations and 12-dimensional annotations are highlighted with purple and yellow, respectively. 
     (b) The Expert head is a sparsely connected MLP, with each node being supervised.}
    \label{fig:model-overview}
  % \vspace{-2em}
  % \includegraphics[width=0.48\textwidth]{figures/model-overview.pdf}
  % % \vspace{-12pt}
  % \caption{\small 
  % \textbf{Our HumanAesExpert, compared to existing state-of-the-art methods, shows exceptional improvements. }$\uparrow$ indicates that larger values are better, while $\downarrow$ signifies the opposite.}
  % \label{fig:leida}
  \vspace{-24pt}
\end{wrapfigure}

\noindent \textbf{Regression Head. }Denoted as $H_{reg}$ \cite{videoscore2024he,lift2024wang}, directly obtain scores $S_{reg}=H_{reg}(h)$ via the last token (i.e. the class token) and use Mean Squared Error (MSE) loss to supervise the model learning, as shown in the following equation:
\begin{equation}
    \mathcal{L}_{reg}=\frac{1}{N}\sum_{n=1}^N(\overline{y}-S_{reg})^2,
\end{equation}
\noindent where $N$ is the number of assessment dimensions, and $\overline{y}$ is the ground truth scores.

\noindent \textbf{Expert Head. }Our Expert head $H_{exp}$ uses a sparsely connected MLP to replicate the relationships between scoring dimensions.
This design is based on the aesthetic evaluation system accumulated by experts over a long period. It is a simple and effective structure suitable for most scenarios.
Specifically, the first layer is a linear layer that directly derives the 9 smallest dimensional scores, depicted in Fig.~\ref{fig:model-overview} (b). 
The general appearance aesthetic score and the facial aesthetic score are obtained from their corresponding smallest dimensional scores using two Feed-Forward Networks (FFNs). Finally, another FFN is used to integrate the environment score with the two parent dimensional scores to achieve the overall aesthetic score. We apply MSE loss to each node to ensure the expected specific score learning. The Expert head can be used to generate 12-dimensional scores $score_1,...,score_{12}=H_{exp}(h)$.

\noindent \textbf{MetaVoter.} To balance the contributions of multiple heads, an additional MLP with batch normalization and ReLU activation functions, termed the Metavoter $V$, is trained to aggregate scores from the three heads for the final prediction: $y_{final}=V(S_{LM}^{\prime},S_{reg},S_{exp})$, where $S_{LM}^{\prime}$ is the normalized $S_{LM}$ using Min-Max normalization, and $S_{exp}=score_{12}$. Mean Absolute Error (MAE) loss is used to supervise the learning of MetaVoter $V$, formulated as follows:

\begin{equation}
\label{eq:metavoter}
\mathcal{L}_{MetaVoter} = \frac{1}{N}\sum_{n=1}^N|\overline{y}-V(S_{LM}^{\prime},S_{reg},S_{exp})|,
\end{equation}

\noindent where $N$ is the batch size.

\noindent \textbf{Training.} As illustrated in Fig.~\ref{fig:model-overview} (a). Our method employs a two-stage training. \textbf{In the first stage}, we fine-tune the VLM and train the newly added Regression and Expert heads. Similar to our data, our training includes two types: (1) Only overall annotation: The VLM processes the question, while the image encoder handles the human image from HumanBeauty-58K. The LM head and Regression head are optimized using the overall aesthetics scores and rating levels. (2) 12-dimensional annotations: The VLM receives a conditional question, and the image encoder processes the human images from HumanBeauty-50K. The LM head and Expert head are trained on the 12-dimensional rating levels and corresponding scores. We design a switch between the Regression head and Expert head to control gradient flow and prevent cross-domain conflicts. Our loss function is formulated as follows:

\begin{equation}
\mathcal{L}_{LLM}=\left\{
\begin{array}{rcl}
\mathcal{L}_{Entropy}+\lambda \cdot \mathcal{L}_{reg} & & {f=0}\\
\mathcal{L}_{Entropy}+\mu \cdot \mathcal{L}_{exp} & & {f=1}
\end{array} \right.
,
\end{equation}

\noindent where $\lambda$, $\mu$ are loss balancing terms, $\mathcal{L}_{exp}$ represents the same MSE loss as $\mathcal{L}_{reg}$, and $f$ indicates the data case type.
\textbf{In the second stage}, we freeze all model components except for the MetaVoter and optimize the MetaVoter using the scores from the three heads according to Eq.~\ref{eq:metavoter}.

\section{Experiments}

% \begin{figure}[tb]
%     \centering
%     \includegraphics[width=\linewidth]{figures/4figures.pdf}
%     \caption{Statistical Analysis and Train-Test Split.}
%     \label{fig:statistical-analysis}
% \end{figure}

\subsection{Statistical Analysis and Experimental Setup}
\label{sec:analysis}

\begin{figure*}[ht]
  \centering
  \vspace{-6pt}
  \includegraphics[width=1\textwidth]{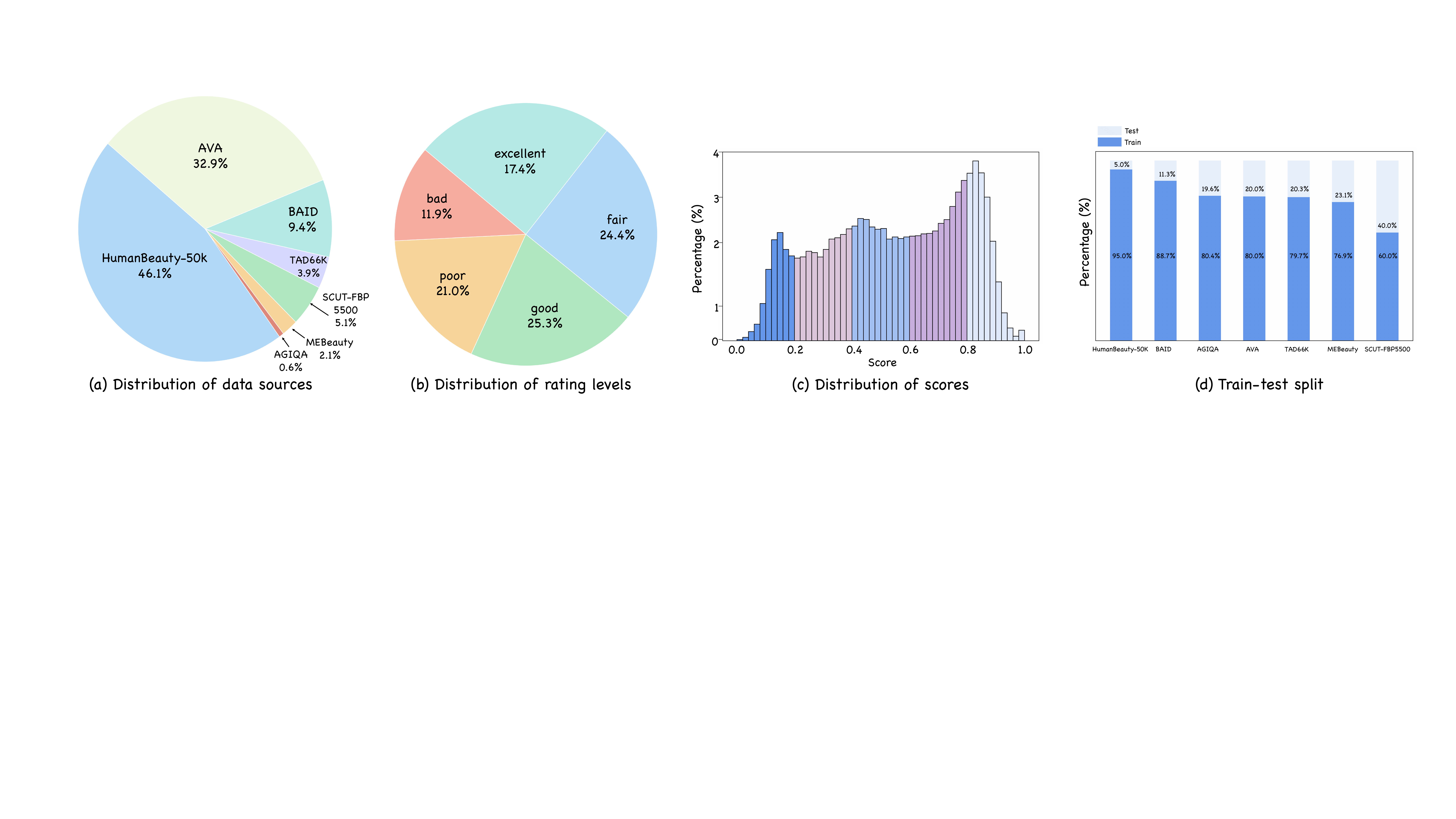}
  % \vspace{-12pt}
  \caption{\small 
  \textbf{Statistical Analysis and Train-Test Split.}}
  \label{fig:statistical-analysis}
  \vspace{-6pt}
\end{figure*}

\textbf{Statistical Analysis. }We present the distribution of our dataset in Fig.~\ref{fig:statistical-analysis} (a), showing that natural images constitute the majority, with artistic images accounting for 9.4\% and AIGC images comprising 0.6\%. This distribution aligns with practical application scenarios of HIAA.
In Fig.~\ref{fig:statistical-analysis} (b) and (c), we illustrate the distributions of rating levels and scores.
The balanced rating levels and the fact that no 0.02 score interval exceeds 4\% demonstrate the diversity of our data and the effectiveness of the annotation process. Additional details are provided in the \textit{Appendix}.

\noindent\textbf{Train-Test Split. }To ensure the balance of the test set, we split the data from the subsets of HumanBeauty by reasonable proportions, shown in Fig.~\ref{fig:statistical-analysis} (d). The training set comprises 94,099 human images, while the test set contains 14,487 human images.

% \noindent \textbf{Implementation Details.} We provide two model variants: 1) HumanAesExpert-1B, consisting of Qwen2-0.5B \cite{qwen22024yang} and InternViT‑300M \cite{internvl22024chen}, and 2) HumanAesExpert-8B, consisting of internlm2-7b and InternViT‑300M. VLM fine-tuning is conducted using the SWIFT framework \cite{swift2024zhao} with LoRA \cite{lora2021hu} on V100 GPUs, while newly added modules are initialized and trained from scratch. To prevent overfitting, the epoch of the first training stage is set to 1, while MetaVoter is trained for 10 epochs in the second stage. 

\noindent \textbf{Implementation Details. }We provide two model variants: 1) HumanAesExpert-1B, consisting of Qwen2-0.5B \cite{qwen22024yang} and InternViT‑300M \cite{internvl22024chen}. 2) HumanAesExpert-8B, consisting of InternLM2‑7B and InternViT‑300M. VLM fine-tuning is conducted by the SWIFT framework \cite{swift2024zhao} with LoRA \cite{lora2021hu} on V100 GPUs, while newly added modules are initialized and trained from scratch. To prevent overfitting, the epoch of the first training stage is set to 1, while MetaVoter is trained for 10 epochs in the second stage. Additional details are provided in the \textit{Appendix}.

\noindent \textbf{Evaluation metrics.} We evaluate the score prediction capabilities of models using several metrics: MSE and MAE measure the distance between the model's predicted scores and the ground truth (GT) scores, reflecting alignment with human aesthetics. Pearson Linear Correlation Coefficient (PLCC), Spearman Rank Correlation Coefficient (SRCC), and Kendall Rank Correlation Coefficient (KRCC) assess the correlation between the model's predictions and the GT scores. Accuracy, mean Precision, mean Recall, and mean F1 Score compare the model's predicted rating levels with those mapped from the GT scores.

\subsection{Quantitative and Qualitative Evaluation}
\label{sec:experiment}

\begin{table*}[tb]
\centering
\caption{\textbf{Comparison with SOTA methods.} $\uparrow$ represents the larger is the better, while $\downarrow$ represents smaller is the better. The upper part of the table lists traditional IAA methods, which use CNN or ViT. The lower part of the table consists of VLM methods. \textbf{Bold} indicates the best, and \underline{underline} indicates the second best.}%
\renewcommand\arraystretch{1.2}
\resizebox{1\textwidth}{!}
{
\begin{tabular}{lcccccccccc}
\toprule
Method                                             & Publication                      & MSE $\downarrow$   & MAE $\downarrow$   & PLCC $\uparrow$    & SRCC $\uparrow$    & KRCC $\uparrow$    & Accuracy $\uparrow$ & Precision $\uparrow$ & Recall $\uparrow$  & F1 Score $\uparrow$ \\ \midrule
TANet \cite{TAD66k2022he}                          & \multicolumn{1}{c|}{IJCAI2022}   & 0.2684           & 0.4905           & 0.3799           & 0.4240           & 0.2820           & 0.1576            & 0.1667             & 0.2052           & 0.1224            \\
CLIPIQA+ \cite{clipiqa2023wang}                    & \multicolumn{1}{c|}{AAAI2023}    & 0.2207           & 0.4259           & 0.2999           & 0.3112           & 0.2117           & 0.2954            & 0.2272             & 0.2164           & 0.1666            \\
VILA \cite{vila2023ke}                             & \multicolumn{1}{c|}{CVPR2023}    & 0.2844           & 0.5031           & 0.2521           & 0.3055           & 0.2139           & 0.1666            & 0.1605             & 0.1864           & 0.1352            \\
EAT \cite{eat2023he}                               & \multicolumn{1}{c|}{ACM MM2023}  & 0.2803           & 0.5036           & 0.3533           & 0.4073           & 0.2722           & 0.1410            & 0.2027             & 0.1900           & 0.1174            \\
QualiCLIP \cite{qualiCLIP2024agnolucci}            & \multicolumn{1}{c|}{ArXiv2403}   & 0.2575           & 0.4409           & 0.2877           & 0.2146           & 0.1435           & 0.3296            & 0.2394             & 0.2270           & 0.2195            \\ \midrule
OneAlign-8B \cite{qalign2023wu}                    & \multicolumn{1}{c|}{ICML2024}    & 0.3000           & 0.5117           & 0.2626           & 0.3051           & 0.2168           & 0.1569            & 0.1515             & 0.1502           & 0.1280            \\
QInstruct-7B \cite{qinstruct2024wu}                & \multicolumn{1}{c|}{CVPR2024}    & 0.2367           & 0.4290           & 0.2125           & 0.2783           & 0.1841           & 0.3270            & 0.2616             & 0.2328           & 0.2138            \\
LLaVA1.6-7B \cite{llava162024liu}                  & \multicolumn{1}{c|}{Blog2401}    & 0.2538           & 0.4239           & 0.2272           & 0.1312           & 0.1025           & 0.2813            & 0.1712             & 0.1820           & 0.1590            \\
Phi3.5-4B \cite{phi2024abdin}                      & \multicolumn{1}{c|}{ArXiv2404}   & 0.2723           & 0.4398           & 0.1321           & 0.1937           & 0.1452           & 0.2387            & 0.2184             & 0.1948           & 0.1651            \\
AesExpert-7B \cite{aesexpert2024huang}             & \multicolumn{1}{c|}{ACM MM2024}  & 0.3947           & 0.5674           & 0.0008           & 0.0033           & 0.0021           & 0.2622            & 0.2319             & 0.1830           & 0.1505            \\
Glm-4v-9b \cite{chatglm2024glm}                    & \multicolumn{1}{c|}{ArXiv2406}   & 0.2965           & 0.4785           & 0.0843           & 0.0329           & 0.0281           & 0.2133            & 0.1891             & 0.2114           & 0.1450            \\
InternVL2-1B \cite{internvl22024chen}              & \multicolumn{1}{c|}{Blog2407}    & 0.3985           & 0.5716           & 0.0001           & 0.0221           & 0.0179           & 0.2779            & 0.1970              & 0.1972           & 0.1916            \\
InternVL2-8B \cite{internvl22024chen}              & \multicolumn{1}{c|}{Blog2407}    & 0.3324           & 0.5006           & 0.0001           & 0.0229           & 0.0157           & 0.2809            & 0.2015             & 0.2014           & 0.1958            \\
LLaVAov-1B \cite{llavaov2024li}                    & \multicolumn{1}{c|}{ArXiv2408}   & 0.2448           & 0.4181           & 0.0201           & 0.0416           & 0.0339           & 0.3413            & 0.0948             & 0.1987           & 0.1074            \\
MiniCPM2.6-8B \cite{minicpm262024yao}              & \multicolumn{1}{c|}{ArXiv2408}   & 0.2656           & 0.4348           & 0.0541           & 0.0859           & 0.0626           & 0.3167            & 0.1851             & 0.1968           & 0.1745            \\
Qwen2VL-7B \cite{qwen2vl2024wang}                  & \multicolumn{1}{c|}{ArXiv2409}   & 0.2383           & 0.4546           & 0.0323           & 0.0708           & 0.0551           & 0.2626            & 0.1851             & 0.1840           & 0.1706            \\
Llama3.2-11B \cite{llama322024meta}                & \multicolumn{1}{c|}{Blog2409}    & 0.3165           & 0.4429           & 0.0791           & 0.1291           & 0.0916           & 0.3160            & 0.2250             & 0.2208           & 0.2061            \\
HumanVLM-8B \cite{humanvlm2024dai}                 & \multicolumn{1}{c|}{ArXiv2411}   & 0.3359           & 0.4958           & 0.1873           & 0.1054           & 0.0844           & 0.2818            & 0.2189             & 0.2196           & 0.2109            \\
\rowcolor[HTML]{E4F1F8} \textbf{HumanAesExpert‑1B} & \multicolumn{1}{c|}{\textbf{--}} & \underline{0.1194} & \underline{0.2906} & \underline{0.8539} & \underline{0.7090} & \underline{0.5286} & \underline{0.5221}  & \underline{0.5195}   & \underline{0.5273} & \underline{0.5016}  \\
\rowcolor[HTML]{CBE5F5} \textbf{HumanAesExpert‑8B} & \multicolumn{1}{c|}{\textbf{--}} & \textbf{0.1141}  & \textbf{0.2849}  & \textbf{0.8632}  & \textbf{0.7259}  & \textbf{0.5459}  & \textbf{0.5478}   & \textbf{0.5707}    & \textbf{0.5652}  & \textbf{0.5334}   \\ \bottomrule 
\end{tabular}
}
\label{tab:compare-results}
\end{table*}

\begin{table*}[tb]
\centering
\caption{\textbf{Comparison on fine-grained HIAA}, where MAE are reported, and ``G-A Aesthetic'' denotes general appearance aesthetic.}%
\renewcommand\arraystretch{1.1}
\resizebox{1\textwidth}{!}{
\begin{tabular}{l|ccccc|c}
\toprule
Method                                             & Facial Brightness & Facial Feature Clarity & Facial Skin Tone                       & Facial Structure                                                         & Facial Contour Clarity & \color[HTML]{4169E1} Facial Aesthetic        \\ \midrule
OneAlign-8B \cite{qalign2023wu}                                       & 0.7501          & 0.4882                & 0.5995                               & 0.4886                                                                 & 0.3880               & 0.3110                                       \\
QInstruct-7B \cite{qinstruct2024wu}                                       & 0.4324          & 0.6597                & 0.4582                               & 0.4576                                                                 & 0.6168               & 0.4482                                       \\
LLaVA1.6-7B \cite{llava162024liu}                                        & 0.5781          & 0.6589                & 0.4701                               & 0.5735                                                                 & 0.5688               & 0.5095                                       \\
HumanVLM-8B \cite{humanvlm2024dai}                                        & 0.7830          & 0.7380                & 0.5775                               & 0.7204                                                                 & 0.7184               & 0.5916                                       \\
\rowcolor[HTML]{E4F1F8} \textbf{HumanAesExpert‑1B} & \underline{0.3209}          & \underline{0.3943}                & \underline{0.3949}                               & \underline{0.3718}                                                                 & \underline{0.3415}               & \underline{0.2920}                                       \\
\rowcolor[HTML]{CBE5F5} \textbf{HumanAesExpert‑8B} & \textbf{0.3028} & \textbf{0.3113}       & \textbf{0.3242}                      & \textbf{0.3413}                                                        & \textbf{0.3101}      & \textbf{0.2412}                              \\ \midrule
Method                                             & Outfit          & Body Shape            & \multicolumn{1}{c|}{Looks}           & \multicolumn{1}{c|}{\color[HTML]{4169E1} G-A Aesthetic} & Environment          & \color[HTML]{1E90FF} Overall Aesthetic \\ \midrule
OneAlign-8B \cite{qalign2023wu}                                       & 0.4153          & 0.3506                & \multicolumn{1}{c|}{0.3232}          & \multicolumn{1}{c|}{0.3526}                                            & 0.3367               & 0.5117                                       \\
QInstruct-7B \cite{qinstruct2024wu}                                      & 0.2996          & 0.3384                & \multicolumn{1}{c|}{0.4228}          & \multicolumn{1}{c|}{0.4054}                                            & 0.5082               & 0.4290                                       \\
LLaVA1.6-7B \cite{llava162024liu}                                       & 0.4320          & 0.5043                & \multicolumn{1}{c|}{0.4765}          & \multicolumn{1}{c|}{0.4268}                                            & 0.5409               & 0.4239                                       \\
HumanVLM-8B \cite{humanvlm2024dai}                                       & 0.5030          & 0.6538                & \multicolumn{1}{c|}{0.5193}          & \multicolumn{1}{c|}{0.5365}                                            & 0.5844               & 0.4958                                       \\
\rowcolor[HTML]{E4F1F8} \textbf{HumanAesExpert‑1B} & \underline{0.2918}          & \underline{0.2728}                & \multicolumn{1}{c|}{\underline{0.2512}}          & \multicolumn{1}{c|}{\underline{0.2986}}                                            & \underline{0.2910}               & \underline{0.2906}                                       \\
\rowcolor[HTML]{CBE5F5} \textbf{HumanAesExpert‑8B} & \textbf{0.1803} & \textbf{0.2704}       & \multicolumn{1}{c|}{\textbf{0.2471}} & \multicolumn{1}{c|}{\textbf{0.2671}}                                   & \textbf{0.2708}      & \textbf{0.2849}                              \\ \bottomrule
\end{tabular}
}
\label{tab:Sub-dimension}
\end{table*}

\noindent \textbf{Quantitative Evaluation. }We conduct a quantitative comparison of overall HIAA and fine-grained HIAA against existing SOTA methods on our test set.
For overall HIAA, we compare with the latest open-source methods as baselines from two main categories: traditional CNN and ViT models \cite{TAD66k2022he,clipiqa2023wang,vila2023ke,eat2023he,qualiCLIP2024agnolucci}, and VLM-based methods \cite{qalign2023wu,qinstruct2024wu,llava162024liu,phi2024abdin,aesexpert2024huang,chatglm2024glm,internvl22024chen,llavaov2024li,minicpm262024yao,qwen2vl2024wang,llama322024meta,humanvlm2024dai}. 
The results, as shown in Tab.~\ref{tab:compare-results}, demonstrate that our HumanAesExpert-8B and HumanAesExpert-1B achieve the optimal and suboptimal performance across all metrics.
Concretely, our 1B and 8B models outperform TANet 
\cite{TAD66k2022he} (the best among other methods), achieving 124\% and 127\% gains in PLCC, 74\% and 78\% in SRCC, and 87\% and 94\% in KRCC, respectively. 
For fine-grained HIAA, we select the top four methods in Tab.~\ref{tab:compare-results} and assess them across our 12 aesthetic dimensions. 
We report the MAE metric in Tab.~\ref{tab:Sub-dimension}, which measures the distance between predicted scores and GT, providing more intuitive evidence of alignment with human aesthetic judgment.
\begin{wraptable}{r}{0.5\textwidth}
    \centering
    \vspace{-12pt}
    \caption{\small \textbf{Zero-shot comparisons.}}%
    \vspace{-4pt}
    \renewcommand\arraystretch{1.2}
    \resizebox{\linewidth}{!}{
    \begin{tabular}{l|lll|lll}
    \toprule
    \multicolumn{1}{c|}{\multirow{2}{*}{Method}} & \multicolumn{3}{c|}{APDDv2 \cite{jin2024apddv2}} & \multicolumn{3}{c}{LAPIS \cite{maerten2025lapis}} \\ \cmidrule(l){2-7} 
    \multicolumn{1}{c|}{}                        & PLCC$\uparrow$    & SRCC$\uparrow$    & KRCC$\uparrow$    & PLCC$\uparrow$    & SRCC$\uparrow$   & KRCC$\uparrow$   \\ \midrule
    OneAlign-8B \cite{qalign2023wu}                                 & \underline{0.5302}  & \underline{0.4411}  & \underline{0.3083}  & 0.2172  & 0.2015 & 0.1369 \\
    QInstruct-7B \cite{qinstruct2024wu}                                & 0.2326  & 0.1862  & 0.1224  & 0.0951  & 0.1215 & 0.0806 \\
    LLaVA1.6-7B \cite{llava162024liu}                                 & 0.4767  & 0.4138  & 0.3375  & 0.1265  & 0.1458 & 0.1194 \\
    HumanVLM-8B \cite{humanvlm2024dai}                                 & 0.2116  & 0.2036  & 0.1659  & 0.3445  & 0.3294 & 0.2697 \\
    \rowcolor[HTML]{E4F1F8} HumanAesExpert-1B                            & 0.5198  & \underline{0.4411}  & 0.3057  & \underline{0.4048}  & \underline{0.4221} & \underline{0.2917} \\
    \rowcolor[HTML]{CBE5F5} HumanAesExpert-8B                            & \textbf{0.7280}  & \textbf{0.5564}  & \textbf{0.392}   & \textbf{0.6544}  & \textbf{0.6254} & \textbf{0.4362} \\ \bottomrule
    \end{tabular}
    }
    \vspace{-10pt}
\label{tab:zeroshot}
\end{wraptable}
Our models consistently achieve the best and second-best results across all sub-dimensions. 
Specifically, compared to QInstruct-7B \cite{qinstruct2024wu}, our 1B and 8B models reduce MAE by 26\% and 30\% on the face brightness sub-dimension and by 19\% and 25\% on the face structure sub-dimension, respectively. Similarly, compared to OneAlign-8B \cite{qalign2023wu}, our models achieve reductions of 22\% and 23\% on the looks sub-dimension. 
These results support the proposed Expert head in achieving fine-grained HIAA learning via a hierarchical sparsely connected network structure. 
In summary, the exceptional improvements of our model in both overall and fine-grained HIAA tasks demonstrate the effectiveness of our holistic framework.
To enable fair comparisons, we apply the data filtration again to both APDDv2 \cite{jin2024apddv2} (obtaining 907 images from 10,022 images) and the LAPIS \cite{maerten2025lapis} test set (307 images from 2,345 images) for zero-shot evaluations, as shown in Tab.~\ref{tab:zeroshot}. Our 8B model still performs the best in zero-shot settings. Please refer to \textit{Appendix} for more details.

\begin{figure*}[ht]
  \centering
  \includegraphics[width=1\textwidth]{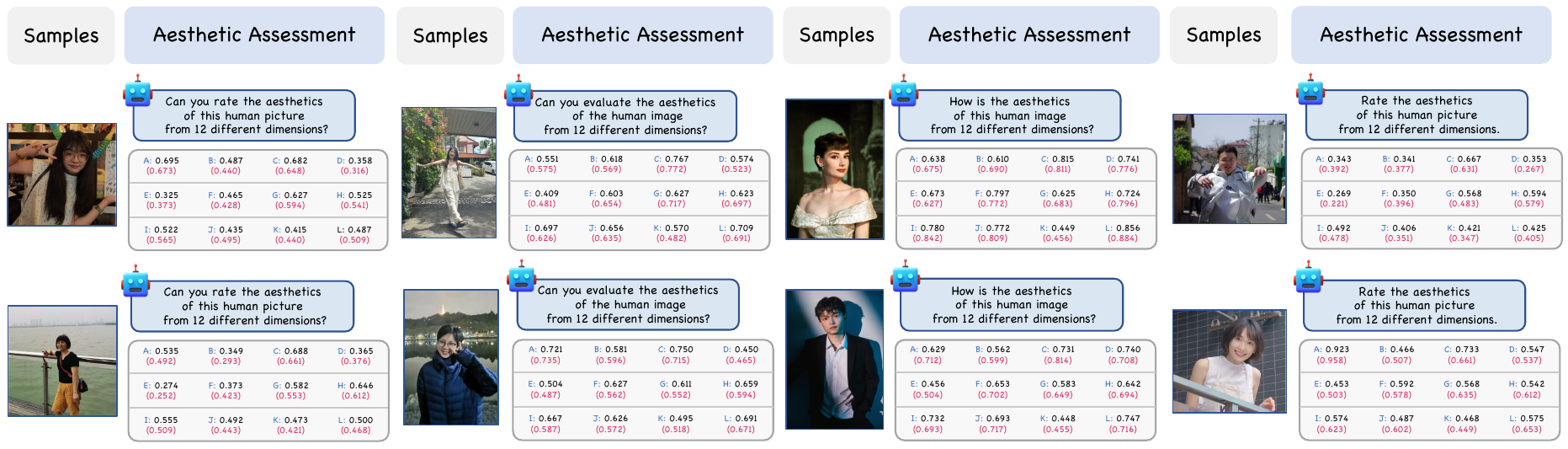}
  \caption{\textbf{The visualization results of our model}, where ``( )'' indicate the Ground Truth scores. From A to L, they respectively represent facial brightness, facial feature clarity, facial skin tone, facial structure, facial contour clarity, facial aesthetic, outfit, body shape, looks, general appearance aesthetic, environment and overall aesthetic scores.}
  \label{fig:case}
\end{figure*}

\noindent \textbf{Qualitative Evaluation.} 
We design a qualitative evaluation by simulating the model usage process. We ask our model to perform aesthetic evaluations on human images across 12 dimensions and directly compare the outputs with the ground truth. 
As shown in Fig.~\ref{fig:case}, the evaluation results of our model closely match human aesthetic annotations, indicating the high aesthetic consistency with humans'.
In the first image, due to hair occlusion obscuring the facial contour, the model assigns a lower score to this dimension. These observations demonstrate the model's capability in fine-grained HIAA. In the fourth image, a notably low overall aesthetic score is observed, attributable to the equally low scores in both general appearance aesthetics and facial aesthetics. Further analysis reveals deficiencies in facial brightness, clarity of facial features, and contour definition. This aligns with our intuitive observation: the facial features of the person appear less distinct due to his heavier build. By systematically tracing back from the total score, we identify the underlying factors contributing to the low evaluation, demonstrating that our established assessment standard is inherently attributable.

% \begin{wraptable}{r}{0.5\textwidth}
%   \centering
%   \begin{tabular}{|c|c|}
%     \hline
%     列1 & 列2 \\
%     \hline
%     数据1 & 数据2 \\
%     数据3 & 数据4 \\
%     \hline
%   \end{tabular}
%   \caption{这是一个示例表格}
% \end{wraptable}

\subsection{Ablation Study}
\label{sec:ablation}

In the ablation study, we randomly selected 138 images from each sub-dataset in the test set to create a balanced quick test set.
Inspired by VideoScore \cite{videoscore2024he}, we use InternVL2-1B and InternVL2-8B \cite{internvl22024chen}, fine-tuned with only the Regression head, as our baselines. As shown in Tab.~\ref{tab:ablation-study}, when we add the LM head, the model performance nearly doubles since the LM head guides the learning of all tokens. 
\begin{wraptable}{r}{0.5\textwidth}
    \Huge
    \centering
    \vspace{-8pt}
    \caption{\small \textbf{Ablation Study on Our Proposed Modules.}}%
    \vspace{-2pt}
    \renewcommand\arraystretch{1.8}
    \resizebox{\linewidth}{!}{
    \begin{tabular}{@{}cccc|cccccccc@{}}
    \toprule
    \multicolumn{4}{c|}{\textbf{Model Modules}} &  & \multicolumn{3}{c}{\textbf{HumanAesExpert-1B}}     &  & \multicolumn{3}{c}{\textbf{HumanAesExpert-8B}}      \\ \cmidrule(r){1-4} \cmidrule(lr){6-8} \cmidrule(l){10-12} 
    Reg-Head & LM-Head & Exp-Head & MetaVoter &  & PLCC $\uparrow$            & SRCC $\uparrow$           & KRCC $\uparrow$           & & PLCC $\uparrow$            & SRCC $\uparrow$           & KRCC $\uparrow$            \\ \midrule
        \ding{51}      &         &          &           &  & 0.1841          & 0.1716          & 0.1176         &  & 0.2173          & 0.1802          & 0.1144          \\
         \ding{51}     &     \ding{51}     &          &           &  & 0.4122          & 0.3943          & 0.2666         &  & 0.4809          & 0.454           & 0.3804          \\
           \ding{51}   &     \ding{51}     &     \ding{51}      &           &  & 0.621           & 0.5456          & 0.377          &  & 0.7153          & 0.7075          & 0.5047          \\
          \ding{51}    &      \ding{51}    &   \ding{51}        &      \ding{51}      &  & \textbf{0.7627} & \textbf{0.7483} & \textbf{0.562} &  & \textbf{0.7968} & \textbf{0.7804} & \textbf{0.5965} \\ 
    \bottomrule 
    \end{tabular}
    }
    \vspace{-7pt}
\label{tab:ablation-study}
\end{wraptable}
Adding our proposed Expert head further boosts the model's performance significantly, as the hierarchical neural network design aligns with the evaluation standard, which enhances VLM's understanding of human aesthetics. Finally, incorporating our proposed MetaVoter yields the best results, indicating that deriving the final score from multiple heads is effective.

% \begin{table}[tb]
% \Huge
% \centering
% \caption{Ablation study on our proposed modules.}%
% \renewcommand\arraystretch{1.8}
% \resizebox{\linewidth}{!}
% {
% \begin{tabular}{cccc|cccccccc}
% \toprule
% \multicolumn{4}{c|}{\textbf{Model Modules}} &  & \multicolumn{3}{c}{\textbf{HumanAesExpert-1B}}     &  & \multicolumn{3}{c}{\textbf{HumanAesExpert-8B}}      \\ \cmidrule(r){1-4} \cmidrule(lr){6-8} \cmidrule(l){10-12} 
% Reg-Head & LM-Head & Exp-Head & MetaVoter &  & PLCC \uparrow            & SRCC \uparrow           & KRCC \uparrow           & & PLCC \uparrow            & SRCC \uparrow           & KRCC \uparrow            \\ \midrule
%     \ding{51}      &         &          &           &  & 0.1841          & 0.1716          & 0.1176         &  & 0.2173          & 0.1802          & 0.1144          \\
%      \ding{51}     &     \ding{51}     &          &           &  & 0.4122          & 0.3943          & 0.2666         &  & 0.4809          & 0.454           & 0.3804          \\
%        \ding{51}   &     \ding{51}     &     \ding{51}      &           &  & 0.621           & 0.5456          & 0.377          &  & 0.7153          & 0.7075          & 0.5047          \\
%       \ding{51}    &      \ding{51}    &   \ding{51}        &      \ding{51}      &  & \textbf{0.7627} & \textbf{0.7483} & \textbf{0.562} &  & \textbf{0.7968} & \textbf{0.7804} & \textbf{0.5965} \\ \bottomrule
% \end{tabular}
% }
% \label{tab:ablation-study}
% \end{table}

\section{Conclusion}
In this paper, we introduce the HumanBeauty dataset with the guidance of our 12-dimensional human aesthetic evaluation standard, which contains 108K images with real annotations, and the HumanAesExpert series of models featuring our proposed Expert head and MetaVoter module. Our experiments demonstrate that our methods achieve state-of-the-art performance on this dataset. Nevertheless, our models still underperform on some metrics, reflecting that our dataset is highly challenging and far from being saturated. Furthermore, we validate the effectiveness of our proposed modules. Our dataset, models, and code are open-sourced, potentially establishing our work as a foundation for HIAA.

\bibliographystyle{plain}
\bibliography{main}

\end{document}